\documentclass[conference]{IEEEtran}
\usepackage[colorlinks=true,urlcolor=blue]{hyperref}
\usepackage{graphicx}
\usepackage{amsfonts}
\usepackage{amsmath}

\newcommand{\vvv}[1] { {\bf #1}}
\begin{document}

\title{Multi-objective evolution for 3D RTS Micro}

\author{
  \IEEEauthorblockN{Sushil J. Louis}
  \IEEEauthorblockA{
    Evolutionary Computing Sytems Lab \\
    Dept. of Computer Science and Engineering\\
    University of Nevada\\
    Reno, NV 89557\\
    \url{http://www.cse.unr.edu/~sushil}}
  \and
  \IEEEauthorblockN{Siming Liu}
  \IEEEauthorblockA{
    Evolutionary Computing Sytems Lab \\    
    Dept. of Computer Science and Engineering \\
    University of Nevada\\
    Reno, NV, 89557 \\
    \url{http://www.cse.unr.edu/~simingl}}
}


\IEEEpeerreviewmaketitle

\maketitle

\begin{abstract}
We attack the problem of controlling teams of autonomous units during
skirmishes in real-time strategy games. Earlier work had shown promise
in evolving control algorithm parameters that lead to high performance
team behaviors similar to those favored by good human players in
real-time strategy games like Starcraft. This algorithm specifically
encoded parameterized kiting and fleeing behaviors and the genetic
algorithm evolved these parameter values. In this paper we investigate
using influence maps and potential fields alone to compactly represent
and control real-time team behavior for entities that can maneuver in
three dimensions. A two-objective fitness function that maximizes
damage done and minimizes damage taken guides our multi-objective
evolutionary algorithm. Preliminary results indicate that evolving
friend and enemy unit potential field parameters for distance, weapon
characteristics, and entity health suffice to produce complex, high
performing, three-dimensional, team tactics.
\end{abstract}


\IEEEpeerreviewmaketitle

\section{Introduction}

Real-Time Strategy (RTS) games model decision making in wartime. The
game genre has become a popular research platform for the study of
Computational and Artificial Intelligence (CI and AI). In RTS games,
players need to establish bases, collect resources, and train military
units with the aim of eliminating their opponents. To be good, RTS
game players require a variety of decision making skills~\cite{bwapi,
  ontanon2013survey}. First, the dynamic environment of RTS games
requires real-time planning on several levels - strategic, tactical,
and reactive. Second, players have to make decisions with imperfect
information within the ``fog of war." Third, like poker, there is no
one optimal strategy; players must model their opponent and adapt
their strategies and tactics to the opponent's playing ``style."
Fourth, players must employ spatial and temporal reasoning to exploit
existing terrain and the time-sensitive nature of actions on a
tactical and strategic level.  All of the challenges described above
and their impact on decision making are crucial for winning RTS games
and there is a steep learning curve for novices.  In the video game
industry, Starcraft (SC) and Starcraft 2 (SC2) players earn
significant prize money and points during a series of championship
games worldwide culminating in a World Series Championship at
Blizzcon~\cite{WCS}. Combining the long time horizon of chess, with
the opponent modeling of poker, and the hand-eye coordination
necessary for fast, fine control of large numbers of game units, RTS
games have garnered significant interest within the computational and
artificial intelligence community. After the success of AlphaGo, many
researchers consider RTS games a significant next frontier in the
computational and articifial intelligence
field~\cite{hassabis2016official, ontanon2013survey}.



An RTS player builds an economy that provides resources to generate
fighting units to destroy the opponent. The term ``Macro'' specifies
economy building and choosing the types and numbers of units to
produce. ``Micro'' refers to the fine-grained, nimble, control of
units and unit groups during a skirmish to inflict maximum damage on
the enemy while sustaining minimal damage to friendly units. Most RTS
games only allow maneuvering in two dimensions (2D) with flying units
restricted to a plane and no real three dimensional (3D)
maneuvering (although there are notable exceptions like Homeworld.\footnote{\url{http://www.homeworldremastered.com/}} This paper attacks the
problem of generating good micro for full 3D maneuvering in the
context of RTS games. We believe our approach is extendable to teams
of autonomous real-world entities.

Each RTS game unit type has different properties that determine the
unit's effectiveness in a skirmish. Depending on unit type, a unit's
weapons will have a specific range, a specific amount of armor, and
deal different damage to different opponent unit types. Each unit type
has different movement speed and can take different amounts of
damage. Given this level of complexity, determining the outcome of a
skirmish usually requires a simulation and good RTS micro can win
skirmishes even when outnumbered and unfavourably positioned.

\subsection{Related work}

Good micro for skirmishes in battle or more generally, coordinated
group behavior for adaptive agents, has applications in many fields
from wargaming and modeling industrial agents, to robotics,
simulations, and video games ~\cite{vadakkepat2000evolutionary,ferber1998meta,
  barbuceanu1995cool,jennings1993commitments,jennings1995controlling,
  olfati2007consensus,egerstedt2001formation, reynolds1987flocks,
  chuang2007multi, dasgupta2008multiagent}.  From flocking, swarming,
and other distributed control algorithms~\cite{reynolds1987flocks,
  chuang2007multi} drawn from observing natural behavior to
evolutionary algorithms that tune the parameters of potential fields
that control the movement of autonomous
vehicles~\cite{vadakkepat2000evolutionary}, much research has been
done in the area of group tactics, adaptive behavior, and distributed
AI. In RTS games, there is strong research interest in generating good
micro since this can lead to your winning skirmishes even when your
forces are outnumbered or otherwise disadvantaged. This paper focuses
on an evolutionary computing approach to generating coordinated
autonomous behaviors (micro) for groups of computer game units (or
entities). Specifically, we use an influence map and several potential
fields to represent entity disposition in order to evolve effective
group tactics, or good micro, for controlling a group of entities
battling an opposing group of entities. Good micro maximizes damage to
enemy units while minimizing damage to friendly units.


Within the games community, Yannakakis~\cite{yannakakis2004evolving}
evolved opponent behaviors while Doherty~\cite{doherty2006evolving}
evolved tactical team behavior for teams of agents. Avery used an
evolutionary computing algorithm to generate influence map parameters
that led to effective group tactics for teams of
entities~\cite{avery2009evolving, avery2010coevolving} against a fixed
opponent.

In physics, a potential field is usually a distance
dependent vector field generated by a force. For example, the force of
gravity generates a gravitational field, an attractive force, on objects with mass. This
gravitational force depends on the distance between the objects and on
their mass. Well designed attractive potential fields can keep
friendly units close together while repulsive potential fields enable
keeping your distance from enemies. Combinations (usually linear vector sums)
of several attractive and repulsive potential fields can lead to
fairly complex behavior~\cite{braitenberg1986vehicles}. Potential fields were
introduced into robotics as a method for real-time obstacle avoiding
navigation and have been used in many video games for controlling multi-unit group
movement~\cite{khatib1986real, olfati2007consensus, egerstedt2001formation, reynolds1987flocks}. In this context, Hagelback
and Johnson used potential fields to drive micro in ORTS, a research
RTS and partial clone of Starcraft, a very popular commercial
RTS~\cite{Hagelback2008using, Starcraft}.

An influence map structures the world into a 2D or 3D grid and assigns
a value to each grid element or cell.  Very early work used
influence maps for spatial reasoning to evolve a \textit{LagoonCraft}
RTS game player~\cite{milecoevol}. Sweetser worked on an AI player
designed with an influence map and cellular automata, where the
influence map was used to spatially model the game world and help the
AI player's decision making in their RTS game
\textit{EmerGEnt}~\cite{sweetser2005combining}. Bergsma \textit{et
  al.} proposed a game AI architecture which used influence maps for a
turn based strategy game~\cite{bergsma2008adaptive}. Preuss \textit{et
  al.} introduced an influence map based path finding algorithm for
group movement in the RTS game
\textit{Glest}~\cite{preuss2010towards,danielsiek2008intelligent}. Su-Hyung
\textit{et al.} used evolutionary neural networks to evolve non-player
characters' strategies based on the information provided by a layered
influence map algorithm in the RTS game \textit{Conqueror}. Uriarte
\textit{et al.} used influence maps for generating kiting behavior and used this for their StarCraft (A popular RTS game and research platform) player \textit{Nova}~\cite{uriarte2012kiting}.  We define potential fields and influence maps in more detail later in the paper.

To the best of our knowledge, Liu's work on evolving micro bots for
Starcraft Brood Wars is closest to the work reported in this
paper~\cite{liu2016Evolving}. Although potential fields and influence
maps guide unit movement, Liu's approach relies on a hand-coded,
parameterized, control algorithm that defines a space of targeting,
firing, and fleeing behaviors. A genetic algorithm then searches this
space for parameters specifying unit behaviors aiming to maximize
damage inflicted on enemy units while minimizing damage to friendlies
against a fixed opponent. Behaviors like kiting (also known as hit and
run) and opponent encirclement emerge and lead to high performing micro
that can defeat this opponent~\cite{liu2013comparing}.

Our work differs from Liu's in that we use a larger number of
potential fields instead of a hand-coded control algorithm specifying behaviors,
and we generalize movement and maneuvering to 3D. Our preliminary results
show that we can quickly evolve unit specific complex behavior
including kiting, fleeing, and englobing. In our experiments with two
types of units called vultures and zealots with different movement
speeds, weapons damage, and health, we can evolve high performing
micro that leads to winning skirmishes. For example, three vultures
(fragile, fast, longer ranged units) learn to disperse and kite a much
larger number of baseline zealots (robust, relatively slow, short
ranged units) while suffering relatively little or no
damage. Furthermore, zealots then evolve against these high
performing vultures and learn micro that increases vulture damage with
tactics such as englobing or dispersing into smaller groups. This kind
of manual co-evolution points towards our future work where we plan to
co-evolve high performance micro for RTS games.

The next section introduces potential fields, influence maps, and
describes our simulation test-bed. Section~\ref{Representation}
specifies our representation and describes the genetic algorithm used
to evolve potential field and influence map parameters. We then
provide and discuss our results. The last section summarizes our
results and conclusions and points out directions for future work.

\section{Simulation Environment} {\label{Simulation}}

Starcraft, released in 1998 and Starcraft 2, released in 2010 are the
most popular RTS games~\cite{Starcraft}. In the RTS AI research
community, \textit{StarCraft} has gained popularity as a research
platform due to the existence of the StarCraft: Brood War Application
Programming Interface (BWAPI) framework, and the
\textit{AIIDE} and \textit{CIG} StarCraft AI tournaments~\cite{bwapi} which use the BWAPI to compare
Starcraft ``Bots,'' or Starcraft AI players.  Since Starcraft is
essentially 2D and we do not have access to Starcraft's source code,
we cannot use the BWAPI for our research. Instead we use FastEcslent,
our open source, 3D capable, modular, RTS game
environment. FastEcslent was developed at the Evolutionary Computing
System's Lab for computational intelligence research in games,
human-robot and human-computer interaction, and other
applications~\cite{FastEcslent}. We note that FastEcslent's graphics
engine runs on a separate thread and can be turned off, a useful
feature for use with evolutionary computing techniques. We modeled
game play in FastEcslent to be similar to StarCraft.

We modeled two units from Starcraft, Vultures and Zealots.  A Vulture
is a somewhat fragile unit with low hit-points but high movement speed
and a ranged weapon. Vultures are effective when outmaneuvering slower
melee units. A Zealot is a melee unit with short attack range and low
movement speed but has high hit-points.
In the rest of this paper, we use the term FVulture and
FZealot to indicate that our units operate in 3D and are the {\bf
  F}lying versions of the Starcraft units.  Since our research focuses
on micro, we disable ``fog of war" in our scenario.
\begin{table}
	\begin{center}
		\footnotesize
		\renewcommand{\arraystretch}{1.3}
		\caption{Unit properties defined in FastEcslent}
		\begin{tabular}{|l|c|c|}
			\hline
			Property         &    Vulture        & Zealot       \\
			\hline
			Hit-points        &     80            &   160        \\
			\hline
			MaxSpeed          &     64            & 40           \\
			\hline
			MaxDamage         & 20                 &  $16\times2$         \\
			\hline
			Weapon's Range             & 256               & 224          \\
			\hline
			Weapon's Cooldown          & 1.1                & 1.24        \\
			\hline
		\end{tabular}
		\label{tab:entityparams}
	\end{center}
\end{table}

We enabled 3D movement by adding maximum ($1000$) and minimum ($0$)
altitudes, as well as a climb rate, $r_c$, of $2$. All units use the
RTS physics used in Starcraft and Starcraft2\cite{Starcraft}.


Noting that a fvulture's faster speed theoretically makes it possible
for a single expertly microed fvulture to kite a group of starcraft's
AI controlled fzealots until all fzealots are eliminated, our
experimental scenarios pitted three fvultures against thirty
fzealots. Our baseline AI consisted of hand tuned potential
field parameters that essentially caused fzealots to move towards and
attack the closest enemy unit.

To evaluate a member of the evolving population, we extracted
potential field parameters, created and initialized a FastEcslent
scenerio with these extracted parameters controlling our player's
micro and played against our baseline AI. At the end of $max$
simulation steps or if all units for one side were destroyed, the
resulting score was returned as this member's fitness. The next
section describes our representation and evaluation function in more
detail.

\section{Representation and Genetic Algorithm} {\label{Representation}}

When we start our FastEcslent scenario, our player moves towards a
target location defined by the lowest value cell in our evolving 3D
influence map (IM). An IM is a 3D grid with values assigned to each
cell by an IM function. An IMFunction is usually specified by two
parameters, a starting influence for the location of the entity and a
maximum range (of this influence) in a 3D game world. The influence linearly
decreases to zero as range increases. In this research, we extend our
IMFunction from using two parameters (starting influence and maximum range) to
five parameters in order to represent more information as shown in
Equation~\ref{IMFunction}.

\subsection{Influence maps}

To calculate any grid-cell value $V_c$, we add the influence from each
of the units within the range $r$ from the cell. $r$ is measured in
number of cells. The influence of a unit at the cell occupied by the unit is computed as
the weighted linear sum below.
\[
I_s = w_1 R_h + w_2 R_c + w_3
\]
where $I_s$ is the starting influence, $R_h$ is the entity's
percentage health, $R_c$ is the entity's percentage cooldown time, and
the $w_i$ for $i = [1..3]$ are evolvable weights. For cells within $r$
cells of the unit, we then compute the decrease per unit $r$ as a fraction of the starting influence.
\[
I_d = I_s  I_f
\]
The influence exerted by a unit at a distance $d$ varying from $0$ to $r$ from the unit location is then
\[
I = I_s - (d  I_d)
\]
So the cell at the unit location receives an influence $I_s$ and the
eight neighboring cells (at distance $1$) receive $I_s - I_d$
influence, while cells at distance $2$ receive $I_s - (2 I_d)$.
Thus the IM value at a particular sum given by the sum of all entities' influence on that cell is
\begin{equation}
\label{IMFunction}
V_c = \sum_{u \in r}(I_s - (d_u I_d))\\
\end{equation}
for all entities $u$ at distance $d_u$ of the cell. The genetic
algorithm evolves $r, I_f, w_1, w_2$ and $w_3$.

This IMFunction not only considers units' positions in the game world
but also includes the hit-points and weapon cooldown of each unit. For
example, an enemy unit with low hit-points or its weapon in cooldown
can, with the right $w_i$, generate a low influence map value. Our
player always select the cell with the minimum value as the target
attack location. Thus for example, our units can move towards and
attack a damaged unit (low hitpoints) that cannot fire until the
weapon cooldown period is over. In essence, changes in the five
parameters ($w1$, $w2$, $w3$, $r$, $\Delta f$) of the IMFunction
specifies a target location for friendly units. ``Move towards the
lowest IM value location attacking enemy units as they come within
range'' is our target selection algorithm. This very simple algorithm
can be contrasted with the more complex target selection and movement
algorithms found in ~\cite{liu2016Evolving}.

\subsection{Potential Fields}

The IM provides a target location to move toward and we use potential
fields to control unit movement to this location. Potential fields are
a type of vector field over space of the form
\begin{equation}{\label{pf}}
f = c d ^ e
\end{equation}
where $f$ is the field strength in the direction of the entity
producing the field, $c$ and $e$ are constants and $d$ is distance
(usually).
Potential fields in robotics, and in games, have traditionally
been used for obstacle avoidance where the target location manifests
an attractive potential field and all obstacles repel the controlled
entity. The vector sum of these potential fields guides
navigation. The direction of the force is in the direction of the
vector difference {\em from} the other unit. $c$ and $e$ are the
evolvable parameters for each of the potential fields in our work.

For specific values of $c$ and $e$, decentralized flocking and
swarming behaviors emerge when large numbers of units independently
use potential fields to maneuver~\cite{AIForGameDevelopers}. Since
good RTS micro takes into account unit health, position, and weapon
state we define and use an attractive and a repulsive potential field
for each of these factors. In addition, we assume that enemy and
friendly units will generate different potential fields.

Specifically, the potential field controlling a unit's behavior is
computed from:
\begin{itemize}
\item The influence map provided target position ( $\vvv{T}$)
\item The potential fields generated by health ($\vvv{H}$), distance  ($\vvv{D}$), and weapon cool-down state ($\vvv{W}$)  of all other units in the game 
\end{itemize}
Thus the potential field acting on a unit at position $\vvv{p}$ is
 the sum of the four potential fields shown below in Equation~\ref{PPFformula}.
\begin{equation}
\label{PPFformula}
\vvv{F}  = \vvv{T} + \vvv{D} + \vvv{H} + \vvv{W} 
\end{equation}

This potential field controls the unit's desired heading and desired
speed.  The desired heading points in the direction of the vector and
the desired speed is proportional to the difference between current
and desired unit heading. Unit speed thus increases (up to a maximum)
if the unit is already pointing in the direction of the potential
field and decreases otherwise. In our game, unit speeds vary from $0$
to a maximum ($s_{max}$).

The $\vvv{D}, \vvv{H}$, and $\vvv{W}$ potential fields are composed from
\begin{itemize}
\item A direction given by the normalized vector difference between
  the current unit's 3D position $\vvv{p}$ and the other unit's 3D
  position ($\vvv{u}$)
  \begin{equation}
    \label{PFDirection}
  \vvv{n} = \frac{(\vvv{u} - \vvv{p})}{ \lvert \vvv{u} - \vvv{p} \rvert}
  \end{equation}
  where $\lvert \vvv{p} - \vvv{u} \rvert$ denotes the length of the
  difference vector.

\item A magnitude given by either the distance to the other unit, the
  health of the other unit, or the cool-down time of the other unit.

\item Whether the other unit is a friend or enemy
\end{itemize}

For example, $\vvv{D}$ is given by
\[
\vvv{D} = \sum_{i \in F} (\, \vvv{n} c_1 d_i^{e_1} - \vvv{n} c_2 d_i^{e_2} )\,
        + \sum_{i \in E} (\, \vvv{n} c_3 d_i^{e_3} - \vvv{n} c_4 d_i^{e_4} )\,
\]
where $d$ is distance, $\vvv{n}$ is given by equation
~\ref{PFDirection}, and where odd subscripts for $c$ and $d$
correspond to attraction and even subscripts correspond to
repulsion. The first summation is over friends ($F$) and the second
over enemies ($E$). In the same way, $\vvv{H}$ is given by
\[
\vvv{H} = \sum_{i \in F} (\, \vvv{n} c_5 h_i^{e_5} - \vvv{n} c_6 h_i^{e_6} )\,
        + \sum_{i \in E} (\, \vvv{n} c_7 h_i^{e_7} - \vvv{n} c_8 h_i^{e_8} )\,
\]
where $h_i$ denotes the health percentage of the $i^{th}$ other unit, and $\vvv{W}$ is
\[
\vvv{W} = \sum_{i \in F} (\, \vvv{n} c_9 w_i^{e_9} - \vvv{n} c_{10} w_i^{e_{10}} )\,
        + \sum_{i \in E} (\, \vvv{n} c_{11} w_i^{e_{11}} - \vvv{n} c_{12} w_i^{e_{12}} )\,
\]
where $w_i$ denotes the percentage time remaining before the weapon on unit $i$ is ready to fire again.
Finally, there is a single attractive potential field exerted by the target location.
\[
\vvv{T} = \vvv{m} c_{13} d_t^{e_{13}}
\]
where $d_t$ is the distance to the target location and $\vvv{m}$ is the direction vector of the target location. 

Thus $13$ potential fields guide unit movement. The $c$ coefficients
have a range of $[-10000..10000]$ and we use $12$ bits for encoding
these parameters. The $e$ exponents range from $[-7..8]$ and we use
$4$ bits for their encoding. We thus need $12 + 4 = 16$ bits per
potential field and $16 \times 13 = 208$ bits for all potential
fields. The influence map parameter $r$ ranges between $0$ to $8$
cells while $\delta, w_1$, and $w_2$ range between $0$ and
$1$. Finally $w_3$ also ranges between $0$ and $8$. This requires $18$
bits for a total binary chromosome length of $208 + 18 = 226$. 

When FastEcslent receives a chromosome, it decodes the binary string
into corresponding parameters. These parameters then control friendly
units as they move in the game world. The fitness of this chromosome
is then computed based on the amount of damage done to enemy forces
and the amount of damage received by friendly forces during a
simulation run in the game world.


\subsection{Fitness Evaluation}

We evaluate individuals in the genetic algorithm's population in
FastEcslent. The decoded chromosome specifies the parameters needed to
compute potential fields and the influence map within
FastEcslent. Every tick of the underlying simulation clock, friendly
units compute potential fields and update their desired headings and
desired speeds. 
If
enemy units come within weapons range of a friendly unit, the friendly
unit targets the {\bf nearest} enemy unit. One difference from
Starcraft, is that all units in FastEcslent can fire in any
direction (oriented firing will be investigated as future work).

With the right set of potential fields and influence maps we expect
friendly units using this simple targeting algorithm to significantly
damage enemy units while themselves taking minimal damage. The simple
fitness function below tries to maximize damage done and minimize damage taken.
\[
fitness = \sum_{enemies} D_e - \sum_{friends} D_f
\]
Where $D_e$ denotes damage to enemy units and $D_f$, damage to
friendly units. This simple fitness function however, tends to
generate fleeing behavior that simply minimizes damage to friendly
units~\cite{liu2016Evolving}. One approach to solving this issue is to
wait for the GA to find some way out of this fleeing behavior, another
approach is to fiddle with the fitness function (for example, give
more weight for damage done) to enable the GA to more easily find a
way out~\cite{liu2016Evolving}. Instead, in this paper, we re-frame
the problem as the problem of exploring the tradeoff between fleeing
and fighting. Evolutionary multi-objective optimization provides
an elegant framework for finding a pareto front of solutions that
explore the tradeoff between damage done and damage received.

Our two objective fitness function is then specified by
\[
fitness = [( \sum_{enemies} D_e ), (1 -  \sum_{friends} D_f )]
\]
where both $D_e$ and $D_f$ are normalized to be between $0$ and $1$
and we are trying to maximize both objectives. We use our own
implementation of Deb's well known non-dominated sort elitist genetic
algorithm NSGA-II as our evolutionary multi-objective optimization
algorithm~\cite{deb2002fast}. The normalized, two-objective fitness
function used within our NSGA-II implementation then produces the
results described in the next section.

\section{Results}

We begin by describing experiments conducted to investigate our
approach to evolving RTS micro. We drew inspiration from Starcraft and
earlier work by Liu using BWAPI and created two unit types similar to
vultures and zealots. Our unit equivalents can fly and are hence named
fvultures and fzealots.  Unit properties were specified earlier in the
paper in Table~\ref{tab:entityparams}. Because fvultures can kite much
larger numbers of fzealots dealing massive damage while receiving
little, our experiments used $3$ fvultures versus $30$ fzealots.


In our initial experiments, we had one clump of three fvultures on the
left against a clump of $30$ fvultures on the right side of the map. A
clump is defined by a center and a radius. All units in a clump are
distributed randomly within a sphere defined by this radius
($400$). Although fvultures learned to do well against the $30$
fzealot clump, their performance degraded significantly when fzealots
were not initially clumped. In addition, when we tried to evolve
fzealot micro against fvultures, we found the same kind of
over-specialized behavior emerging. To reduce overspecialization, we
created three training scenarios (or maps). These scenarios are
defined in terms of clumps and clouds. We have already defined
clumps. A cloud, like a clump is defined by a center and a
radius. However, the units in a cloud are distributed randomly within
$10$ units around the sphere boundary defined by the center and radius
(also $400$). The first scenario consists of a clump of $3$ fvultures
versus a clump of $30$ fzealots. The second, a clump of $3$ fvultures
surrounded by a spherical cloud of $30$ fzealots, and the last
scenario consists of a clump of $30$ fzealots surrounded by a (small)
cloud of $3$ fvultures.

Unit locations within the clump and cloud were generated randomly. Our
evaluation function ran each of these three scenarios and the
objective function values for each objective were averaged over these
three scenarios. Evaluations took longer but the resulting performance
was better and seemed to generalize well.

\subsection{Pareto front evolution}

\begin{figure}[htp] 
  \centerline{
    \includegraphics[width=3.5in]{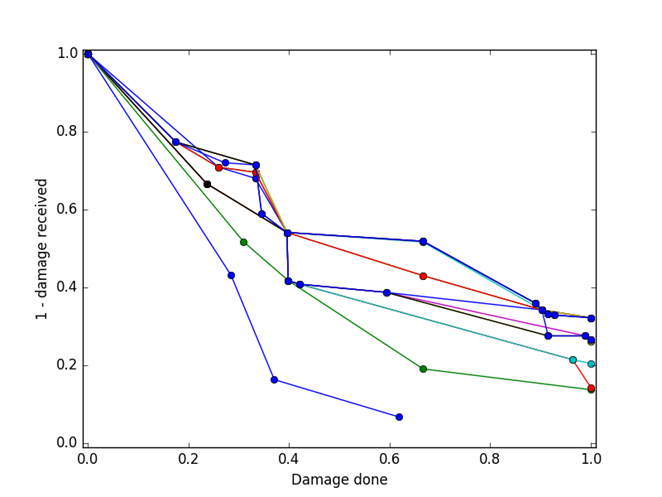}
  }
  \caption{Fvultures evolving against Fzealots}
  {\label{Fvultures}} 
\end{figure}

\begin{figure}[htp] 
  \centerline{
    \includegraphics[width=3.5in]{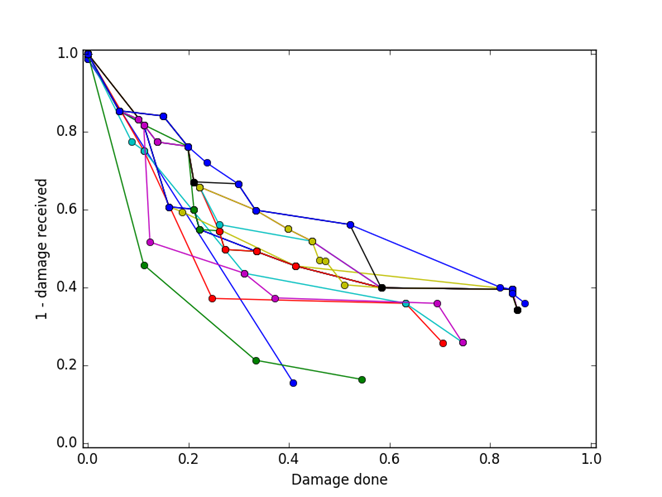}
  }
  \caption{Fzealots evolving against Fvultures}
  {\label{Fzealots}} 
\end{figure}

Figure~\ref{Fvultures} shows the evolution of the pareto front at
intervals of five ($5$) generations for one run of our NSGA-II. We
used a population size of $50$, run for $75$ generations. Simple two
point crossover with a crossover probability of $0.9$ and flip
mutation with a probability of $0.05$ worked well with our
implementation of NSGA-II. The $x$-axis plots the first objective,
damage done to fzealots, and the $y$-axis plots $1 - $ damage
received by fvultures.
Figure~\ref{Fzealots} shows the same information for fzealots evolving against already evolved fvultures.
Figure~\ref{Fvultures} and ~\ref{Fzealots} illustrate how the pareto
front progresses for one run of the multi-objective GA. Broadly
speaking, the pareto front moves towards $(1, 1)$ on the plot making
progress towards maximizing damage done and minimizing damage
received. We evolve micro that can destroy all fzealots but this also
involves taking significant damage as shown by the solutions on the
bottom right of Figure~\ref{Fvultures}. We also see that when battling
fvultures from this area of the plot, fzealots can do significant
damage but never evolve micro to kill all the fvultures during the
single run of the NSGA-II shown in Figure~\ref{Fzealots}.

\begin{figure}[htp] 
  \centerline{
    \includegraphics[width=3.5in]{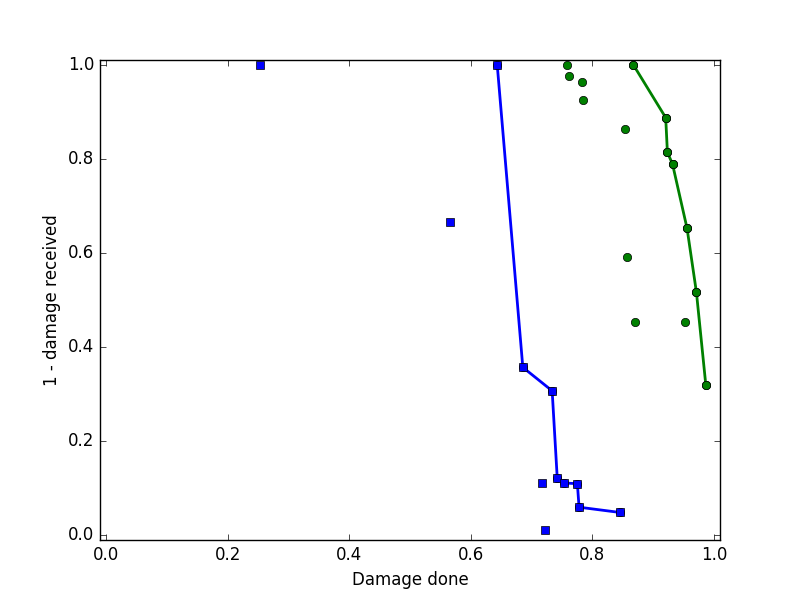}
    }
  \caption{The initial and final pareto front over ten runs for Fvulture micro.}
  {\label{FirstLastFVultures}} 
\end{figure}

In our experiments, we ran the genetic algorithm ten times with
different random seeds and Figure~\ref{FirstLastFVultures} shows the
result. This figure plots the combined pareto front in the first
generation over all ten random seeds versus the combined pareto front
in the last generation over the ten random seeds. That is, we first
did a set union of the pareto fronts in the ten initial randomly
generated populations. The points in this union over all ten runs are
displayed as squares for the initial generation (generation $0$)
points and as circles for the points in the final generation
(generation $74$). Let us label the union of the generation $0$ pareto
fronts as $U_0^{10}$ and the union of the generation $74$ pareto
fronts as $U_{74}^{10}$.  We then compute and plot the pareto front of
this set union - the line on the left. We do the same for the ten
pareto fronts in the last generation - the line on the right. Let
$P^{10}_0$ denote the combined ten run pareto front in the first
generation and $P^{10}_{74}$ denote the combined ten run pareto front
in the last generation; Figure~\ref{FirstLastFVultures} then shows
progress between first and last generation over all ten runs against
an fzealot from the bottom right of Figure~\ref{Fzealots}. Note that
$P_0^{10}$ represents not an average, but the best (non-dominated)
individuals over all ten runs. Individual runs produce initial pareto
fronts that look like the intial front in Figure~\ref{Fvultures}. We
also have videos that show how the fvultures and fzealots behave at
\url{http://www.cse.unr.edu/~sushil/RTSMicro}.
Figure~\ref{FirstLastFVultures} shows that even initial generations
contain a range of acceptable micro behaviors. Since the target
location is near the opponents and exerts a purely attractive
potential field and units fire when in range, fzealots and fvultures
usually come together and fight. As the potential field parametes
evolve, fvultures learn to take less damage while doing more damage to
fzealots. The videos support this observation and also show the range
of micro behaviors evolving. Some can be simply described as
``fleeing,'' while others, even in early generations, defy succint
explanation. Broadly speaking, the micro behaviors being evolved tend
to keep fvultures outside the weapons range of the fzealots while
looking for an opportunity to dart in opportunistically and deliver
damage. Early on, if fvultures are surrounded by zealots as in
training scenario two, the fvultures inflict significant damage but
have not learned to kite and so also suffer significant damage. It is
worthwhile to note that the opponent fzealots tend to favor tight
circling on a plane, sticking close together while the fvultures are
more dispersed, tend to change altitude more frequently, and try to
keep their distance from fzealots as they try to balance their
attractive and repulsive potentials.

In later generations the fvultures have learned potential field
parameter values that lead to kiting (or kiting-like) behavior and
this is reflected in the pareto front moving further to the right
(more damage done to fzealots) and higher (less damage taken by
fvultures). At generation $74$, we have a choice of micro
control tactics provided by the points on the pareto front at
generation $74$. At one extreme, indicated by the point at the very
top on the generation $74$ pareto front $(.87, 1.0)$, all three
fvultures stay alive with no damage while inflicting $87\%$ damage on
the fzealots. $87\%$ damage corresponds to killing $26$ out of the
$30$ fvultures. At the other extreme, denoted by the point on the
bottom right $(.99, .32)$, two fvultures die while killing $99\%$ of
the fzealots. The points in between these extremes represent other
damage done versus damage received tradeoffs and picking which of
these tactics to use depends on the player's in-game needs of the
moment. That our multi-objective approach naturally provides a range
of choices for micro is an elegant side affect of this approach.

\subsection{Random parameter values}

\begin{figure}[htp] 
  \centerline{
    \includegraphics[width=3.5in]{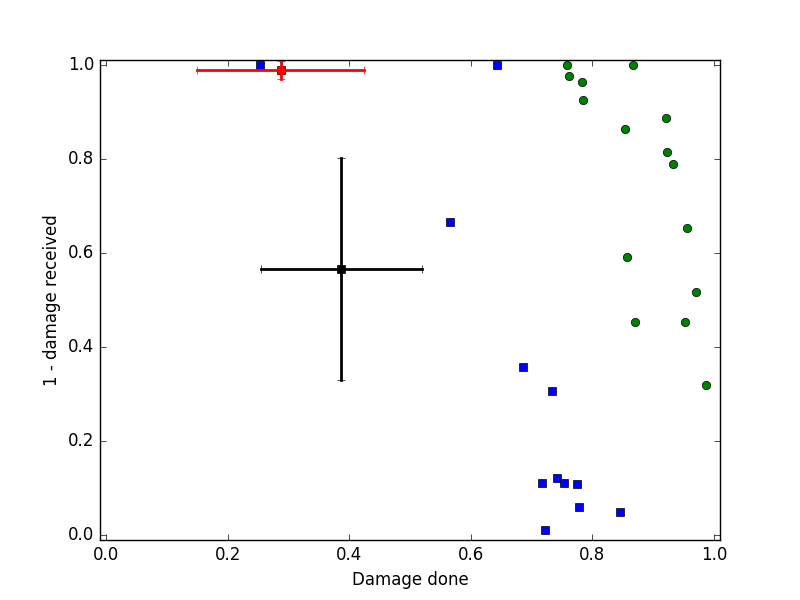}
  }
  \caption{Random chromosome performance on three training scenarios and on $100$ random scenarios}
          {\label{MonteCarlo}}
\end{figure}

We also investigated how random potential field parameter values
perform on our three scenarios and on random scenarios. We generated
random chromosomes that specified these parameter values and evaluated
them on the three training scenarios and on a set of on hundred random
scenarios. In these random scenarios, the starting positions and
orientations of the $30$ fzealots and $3$ fvultures were randomly
spread within a larger clump with radius
$500$. Figure~\ref{MonteCarlo} plots the average objective values and
their standard deviations obtained from evaluting the $3750$ randomly
generated chromosomes on both sets of scenarios. The average objective
value on the three scenarios is average along each objective over
$3750$ evaluations. For the hundred ($100$) random scenarios, the
average is along each objective over $3750 \times 100$ evaluations.
The $3750$ comes from a $50$ population size multiplied by $75$
evolutionary generations equating to the number of evaluations during
one run of the GA. The horizontal and vertical lines on the plot
indicate variation along each objective and are centered at the
average objective values over the $3750$ random chromosomes on the
three training scenarios and the hundred random
scenarios. Specifically, the length of a line equals one standard
deviation and we display half a standard deviation on either side of
the average. We also plot $U_0^{10}$ and $U_{74}^{10}$ for reference.
Figure~\ref{MonteCarlo} shows that the average randomly generated
micro behavior on our three training scenarios has fvultures fleeing
fzealots and doing little damage while much little if any damage. On
the other hand, randomly generated behaviors on random scenarios
average out to be near the center of the pareto plot. They take a
little less damage than they mete out. Both these behaviors can be
explained by the positioning of units in the scenarios. Assuming
random parameter values produce random movement, initial positioning
will play a large part in determing both objectives in the fitness
function. The three training scenarios seperate opposing units and so
fvultures tend to stay seperated from other units and take little damage. On
the random scenarios, randomly positioned fvultures and fzealots moving
randomly will take damage from each other before their random movement
places them out of range and the clump disperses.

\subsection{Random locations}

\begin{figure}[htp] 
  \centerline{
    \includegraphics[width=3.5in]{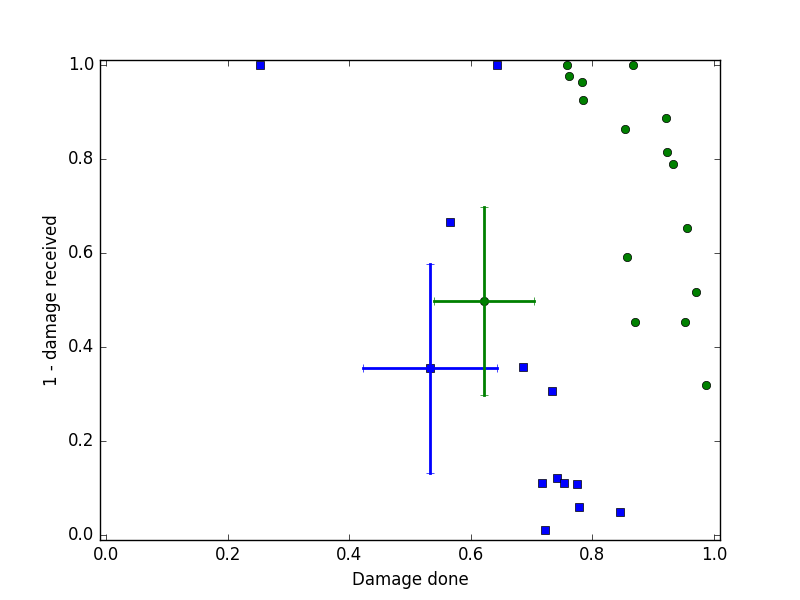}
  }
  \caption{Average performance of $P_0^{10}$ and $P_{74}^{10}$ on a set of $100$ random starting positions and orientations}
  {\label{FirstLastFVulturesRandom}} 
\end{figure}

These results provide evidence that we can use our approach to evolve
a range of good micro against a fixed opponent on a given set of
scenarios. But does the evolved micro generalize to other scenarios?
To investigate this question we also ran the solutions in $P_0^{10}$
and $P_{74}^{10}$ on a set of $100$ random scenarios (as described
above).  The average value of the two objectives over all $100$ random
scenarios and their standard deviations along each objective obtained
by evaluating the individuals in \(P^{10}_{0}\) and \(P^{10}_{74}\) on
these random scenarios is shown in
Figure~\ref{FirstLastFVulturesRandom} along with $U_0^{10}$ and
$U_{74}^{10}$, again for reference.
With arbitrary starting locations and orientations that do not
correspond to training scenarios, we do not expect fvultures to do as
well on average. We also expect wide variation in performance.  As
expected the figure shows that the average performance of the
individuals from $P_0^{10}$ on the $100$ random scenarios is dominated
by the points on the initial pareto front provided by $U_0^{10}$. Let
us label this average performance $R^{100}_{0}$. The average
performance of the individuals from $P_{74}^{10}$ on the $100$ random
scenarios ($R_{74}^{100}$) shows improvement over $R^{100}_{0}$, but
this is still dominated by points on $U_{74}^{10}$. We also see the
existence of significant variation along both objectives and that the standard
deviations along both objectives decreases from $R^{100}_{0}$ to
$R^{100}_{74}$. It is encouraging to see that although performance
degrades on these $100$ random scenarios, individuals from generation
$74$ perform better on average than individuals from the initial
generation on the training scenarios. The change in average
performance from the initial generation to the final generation works
out to approximately three ($3$) more fzealots eliminated and one more
fvulture surviving. Although we were not specifically after robust
solutions, the micro tactics from $R_{74}^{10}$ on average do
eliminate the majority of fzealots, while keeping the majority of
fvultures alive on random never-seen scenarios. The figure provides
evidence that initial unit positioning can significantly affect
skirmish outcome. This is true for initial unit positioning in RTS
games as well.

\section{Conclusion and Future Work}

This paper proposed a new approach to evolving micro for RTS games in
three dimensions for two different units. Using an influence map and
an attractive potential field to determine a target position to
attack, we used attractive and repulsive potential fields for
distance, health, and weapon cooldown to evolve micro behavior for
units in our 3D RTS game arena. Our NSGA-II implementation optimized
micro based on a multi-objective formulation of the fitness function
that maximized damage done and minimized damage taken. The results
show that we can evolve a range of micro behaviors that go from taking
and doing little damage, through micro that balances damage done and
taken, and finally to micro that does maximal damage while taking some
damage. On the three training scenarios, the evolved micro can
eliminate $25$ out of $30$ fzealots while keeping all three ($3$)
fvultures alive.  Other evolved micro (on the pareto front) can
eliminate all fzealots but at the cost of two fvultures. The
multi-objective problem formulation and the NSGA-II lead to evolving
pareto fronts that directly and naturally produce this wide range of
micro choices. The fast, longer ranged, but fragile units (fvultures)
learned micro behavior similar to hit and run (kiting) that is used by
human players in RTS games like Starcraft. Videos of the micro
behavior of these fvultures versus fzealots is available at
\url{http://www.cseunr.edu/~sushil/RTSMicro} and show the range of
complex behavior evolved.

Our genetic algorithm took $3750$ evaluations to evolve the above
micro. We generated $3750$ random chromosomes and evaluated their
fitness on our three training scenarios as well as a hundred random
scenarios. As expected, the average micro behavior of random parameter
values on the three training scenarios was well dominated by the
evolved pareto front on the three training scenarios. Random
chromosomes also led to an average micro performance that balanced
damage done and taken and were explained based on the random
dispositions of units in these scenarios.

Finally, experiments on a hundred random, never before encountered
scenarios show that the potential field values evolved on three
training scenarios used during evolution do not perform badly on these
new random test scenarios. Although we were not aiming for robustness,
micro performance on random scenarios is balanced and shows
improvement over time.

We plan to follow two paths in the future. First, we would like to
investigate how our approach scales with multiple unit types. That is,
we would like to investigate micro for a group composed from multiple
unit types versus another group also composed of multiple unit
types. Second, we plan to investigate the co-evolution of micro using
our new representation and appoach. Our current results show that we
can evolve good micro against a fixed opponent, we would like to use a
multi-objective, co-evolutionary algorithm to co-evolve a range of
micro that is robust against a range of opposition micro.


\bibliographystyle{IEEEtran}
\bibliography{ssci2016}

\end{document}